# End-to-End Autoencoder–MLP Framework for Sepsis Prediction

*Hejiang Cai[1], Di Wu[1*], Ji Xu[2], Xiang Liu[3], Yiziting Zhu[3], Xin Shu[3], Yujie Li[3], Bin Yi[3]*

[1]*College of Computer and Information Science, Southwest University, Chongqing 400715, China*
[2]*State Key Laboratory of Public Big Data, Guizhou University, Guiyang 550025, China.*
[3]*Department of Anesthesiology, Southwest Hospital, Third Military Medical University (Army Medical University), Chongqing, China.*
* *wudi1986@swu.edu.cn*

## Abstract

Sepsis is a life-threatening condition that requires timely detection in intensive care settings. Traditional machine learning approaches, including Naïve Bayes, Support Vector Machine (SVM), Random Forest, and XGBoost, often rely on manual feature engineering and struggle with irregular, incomplete time-series data commonly present in electronic health records. We introduce an end-to-end deep learning framework integrating an unsupervised autoencoder for automatic feature extraction with a multilayer perceptron classifier for binary sepsis risk prediction. To enhance clinical applicability, we implement a customized down-sampling strategy that extracts high information-density segments during training and a non-overlapping dynamic sliding window mechanism for real-time inference. Preprocessed time-series data are represented as fixed-dimension vectors with explicit missingness indicators, mitigating bias and noise. We validate our approach on three ICU cohorts. Our end-to-end model achieves accuracies of 74.6%, 80.6%, and 93.5%, respectively, consistently outperforming traditional machine learning baselines. These results demonstrate the framework's superior robustness, generalizability, and clinical utility for early sepsis detection across heterogeneous ICU environments.

## 1 Introduction

In the era of big data analytics [1–19] and artificial intelligence [20–43], ICU electronic health records generate vast, heterogeneous time-series streams that offer unprecedented opportunities—and challenges—for early warning systems. Sepsis [44-57], a dysregulated host response to infection leading to life-threatening organ dysfunction, accounts for 11 million deaths annually and demands rapid intervention as patient conditions can deteriorate within hours.
Traditional sepsis prediction relies on machine learning (Random Forest, XGBoost, SVM, Naïve Bayes, logistic regression)[44, 46], which—while effective on small, well-curated datasets—suffers from heavy reliance on handcrafted features, propensity to overfit sparse high-dimensional data, and limited transferability across institutions. Inconsistent measurement frequencies and prevalent missing values in EHRs further reduce sample size, introduce bias, and undermine robustness, potentially delaying life-saving alerts.
We propose an end-to-end autoencoder–MLP framework that jointly learns representations and classification. An unsupervised autoencoder [8, 28,58–62] compresses high-dimensional, sparse time-series into compact, denoised latent codes via successive fully connected layers, automatically capturing core physiological patterns. A supervised MLP then models complex decision boundaries based on these latent embeddings. End-to-end training aligns feature extraction with the classification objective, improving generalization across incomplete and heterogeneous datasets.
To ensure clinical applicability, we introduce two preprocessing steps: (1) a customized down-sampling [44] that retains only the last valid observation once feature completeness reaches 80 % per "data block," thereby reducing noise and balancing classes; and (2) a non-overlapping dynamic sliding window [44] that accumulates incoming data until the same 80 % completeness threshold is met, then triggers a single prediction—preserving real-time responsiveness while minimizing redundant alarms.
Validation on publicly available PhysioNet [44] and FHC [44] cohorts demonstrates that our autoencoder–MLP framework consistently outperforms traditional baselines in prediction accuracy, robustness to missing data and class imbalance, and cross-institution generalization—confirming its efficacy and practical potential for early sepsis diagnosis in diverse ICU settings.

## 2. Methodology

To handle irregular, incomplete, and imbalanced ICU time-series, we propose an end-to-end framework: after forward-filling missing vitals and labs, we extract segments with ≥ 80 % completeness using customized down-sampling (training) and dynamic sliding windows (testing). Remaining gaps are zero-filled to create fixed-size vectors marking missingness. These vectors enter an autoencoder that distills latent physiological patterns, then an MLP performs binary sepsis risk prediction. By jointly training feature extraction and classification, we eliminate manual engineering, boost robustness to missing data[63-67] and class imbalance, and enable practical real-time ICU deployment.

*2.1 Data Processing*

All data were processed on a per-patient basis to preserve each individual's time-series integrity and prevent any data leakage between the training and test sets. We extracted samples from the PhysioNet 2019 Challenge (Hospitals A and B) and the FHC dataset[44], yielding a total of 8,337 patient records in the training set and 889 in the test set (see Table 1 for details). For each patient's raw hourly vital-sign and laboratory time series, missing values were first imputed by forward-filling. During training, we applied the customised down-sampling procedure[44]; during testing, we used a non-overlapping dynamic sliding window[44]—both conditioned on an 80 % feature completeness threshold—to ensure every extracted sample contains high-information-density inputs, thereby improving model reliability and predictive performance.

Table 1  Sepsis data distribution

| Dataset | Split | Total Samples | Sepsis Positive | Sepsis Negative |
|---|---|---|---|---|
| PhysioNet A | Training | 4574 | 1843 | 2731 |
|  | Testing | 457 | 131 | 326 |
| PhysioNet B | Training | 3240 | 1307 | 1933 |
|  | Testing | 324 | 93 | 231 |
| FHC | Training | 523 | 211 | 312 |
|  | Testing | 108 | 31 | 77 |

2.1.1 Customised Down-Sampling: During training, each patient's forward-filled hourly series is partitioned into non-overlapping segments by accumulating successive observations until ≥ 80 % of the D features are present. At that point, the final hour's D-dimensional vector—and its sepsis label—are recorded as one training sample, the segment buffer is cleared, and accumulation resumes. This approach condenses irregular ICU records into high-information instances, excludes intervals with excessive missingness, attenuates noise, and mitigates class imbalance by avoiding repeated sampling of extended non-septic periods.

2.1.2 Dynamic Sliding Window: During testing, we adopt a non-overlapping dynamic sliding window analogous to the training down-sampling. Each patient retains an empty buffer that incoming hourly measurements update. Once ≥ 80 % of the D features are populated, the buffer emits its full D-dimensional vector for inference, is cleared (retaining only the patient ID), and a new cycle begins. This guarantees each prediction relies on an information-dense snapshot comparable to training samples, spaces alerts into clinically meaningful intervals, and suppresses redundant alarms. Finally, all vectors are Z-score normalized using training-set means and standard deviations.

*2.2 End-to-End Model*

To address the fragmentation inherent in conventional sepsis-prediction pipelines, we propose a unified end-to-end deep architecture shown in Figure 1, in which latent representation learning and risk classification are co-optimized within a single differentiable computational graph. This holistic formulation obviates ad hoc imputation heuristics and manual feature engineering by enabling the network to intrinsically infer salient temporal patterns from irregular, sparse ICU time-series. Furthermore, by aligning all model parameters directly with the sepsis-classification objective, the framework enhances resilience to pervasive missingness and label imbalance, while streamlining its translation into clinical practice.

An extensive grid search was conducted to optimize the learning rate and training epochs of our end-to-end autoencoder–MLP framework. On the PhysioNet dataset, a learning rate of $7\times10^{-4}$ with 550 epochs produced the highest classification sensitivity and positive predictive value (PPV). Likewise, on the FHC cohort, the combination of a $1\times10^{-3}$ learning rate and 550 epochs maximized both sensitivity and PPV.

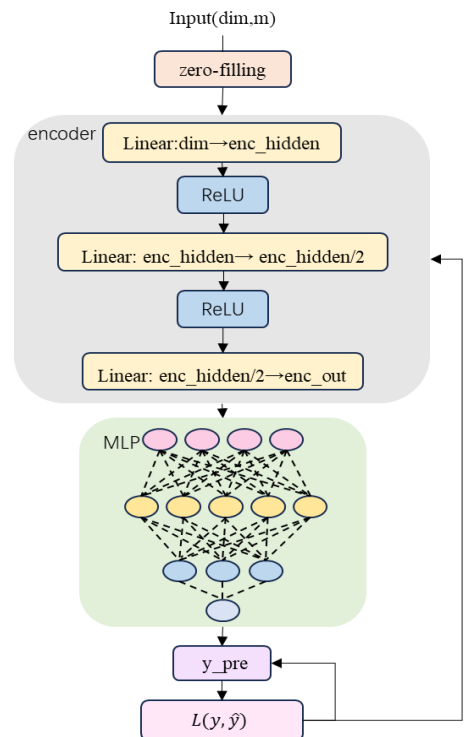

*Figure 1 The end-to-end sepsis classification network*

Each D-dimensional input is passed through a two-stage encoder—fully connected layers of 32 and 16 units—each followed by ReLU activations and 50 % dropout to enforce sparsity and reduce co-adaptation. The resulting 16-dimensional bottleneck distills denoised trajectories of vital signs and laboratory values. A compact MLP head then maps this latent code via a 16→8 hidden layer (also interleaved with ReLU and dropout) into a final linear projection yielding two logits (sepsis vs. non-sepsis). This streamlined design efficiently captures salient physiological patterns and models nonlinear decision boundaries within the learned latent space.

Training proceeds by minimizing a standard cross-entropy loss across all encoder and classifier weights, with stochastic gradient descent facilitating end-to-end convergence. After convergence of the end-to-end training via stochastic gradient descent, we optimize the standard binary cross-entropy loss over the entire encoder–MLP network:

$$L = -\frac{1}{N}\sum_{i=1}^{N}[y_i \log(\hat{y}_i) + (1-y_i)\log(1-\hat{y}_i)] \quad (1)$$

Here, $N$ is the number of training samples, $y_i \in (0,1)$ is the ground-truth label for the iii-th sample (1 for sepsis, 0 for non-sepsis), and $\hat{y}_i$ is the model's predicted probability of sepsis for that sample. This loss directly penalizes erroneous probability estimates, driving gradients through both the encoder and classifier layers so that the learned latent codes become maximally discriminative for the sepsis detection task.

By jointly optimizing representation learning and classification, the model yields latent codes specialized for sepsis detection. In experiments, we benchmark this autoencoder–MLP framework against a tuned gradient-boosting baseline, demonstrating the superiority of end-to-end representation learning on raw, zero-filled ICU feature vectors.

## 3 Results

To quantify the advantages of the proposed end-to-end autoencoder–MLP framework, we performed a comparative assessment against four optimized conventional algorithms (XGBoost, Random Forest, SVM, Naïve Bayes) across three independent ICU cohorts (PhysioNet A, PhysioNet B, FHC), evaluating accuracy, PPV, NPV, sensitivity, and specificity. As detailed in Table 2. The framework yielded on PhysioNet A an accuracy of 74.62 %, PPV of 57.14 %, NPV of 79.83 %, sensitivity of 45.80 %, and specificity of 86.20 %, corresponding to 3–7 % relative improvements over the strongest gradient boosting baseline. Comparable gains were observed on PhysioNet B (accuracy: 80.56 %; PPV: 71.43 %; NPV: 83.07 %; sensitivity: 53.76 %; specificity: 91.34 %) and on the FHC cohort (accuracy: 93.52 %; PPV: 92.86 %; sensitivity: 83.87 %; specificity: 97.40 %). Non parametric Friedman tests rejected the null hypothesis of method equivalence for PPV ($\chi^2 = 10.93$, p = 0.027), NPV ($\chi^2 = 9.87$, p = 0.043), accuracy ($\chi^2 = 11.05$, p = 0.026), and specificity ($\chi^2 = 9.70$, p = 0.046), and post hoc Wilcoxon analyses confirmed statistically significant pairwise improvements (all p < 0.05). Collectively, these findings substantiate that end to end representation learning markedly enhances predictive performance, resilience to missing data and class imbalance, and cross institution generalizability for early sepsis detection.

Table 2 Performance comparison among end-to-end model and traditional machine learning models

| Dataset | Model | PPV(%) | NPV(%) | Accuracy(%) | Sensitivity(%) | Specificity(%) |
|---|---|---|---|---|---|---|
| PhysioNet A | XGBoost | 53.3981 | 78.5311 | 72.8665 | 41.9847 | 85.2761 |
|  | Naïve Bayes | 45.9184 | 76.0446 | 69.5842 | 34.3511 | 83.7423 |
|  | SVM | 45.5556 | 75.4768 | 69.5842 | 31.2977 | 84.9693 |
|  | Random Forest | 53.2609 | 77.5342 | 72.6477 | 37.4046 | 86.8098 |
|  | End-to-End | 57.1429 | 79.8295 | 74.6171 | 45.8015 | 86.1963 |
| PhysioNet B | XGBoost | 65.625 | 80.3846 | 77.4691 | 45.1613 | 90.4762 |
|  | Naïve Bayes | 56.3636 | 76.9517 | 73.4568 | 33.3333 | 89.6104 |
|  | SVM | 45.2174 | 80.3828 | 67.9012 | 55.914 | 72.7273 |
|  | Random Forest | 63.3333 | 79.1667 | 76.2346 | 40.8602 | 90.4762 |
|  | End-to-End | 71.4286 | 83.0709 | 80.5556 | 53.7634 | 91.342 |
| FHC | XGBoost | 81.81 | 94.67 | 90.74 | 87.10 | 92.21 |
|  | Naïve Bayes | 73.5294 | 91.8919 | 86.1111 | 80.6452 | 88.3117 |
|  | SVM | 81.4815 | 88.8889 | 87.037 | 70.9677 | 93.5065 |
|  | Random Forest | 86.6667 | 93.5897 | 91.6667 | 83.871 | 94.8052 |
|  | End-to-End | 92.8571 | 93.75 | 93.5185 | 83.871 | 97.4026 |
| Statistical Analysis | win/loss | 12/0 | 11/1 | 12/0 | 10/2 | 11/1 |
|  | p-value | 0.0273 | 0.0427 | 0.0260 | 0.2288 | 0.0459 |
|  | F-rank | 10.933 | 9.867 | 11.051 | 5.627 | 9.695 |

## 4 Conclusion

The present study introduces an end-to-end autoencoder–MLP architecture that jointly optimizes latent feature representation and binary sepsis risk classification. A customized down-sampling process is applied during training; at inference, a dynamic sliding-window scheme accommodates irregular ICU time-series. On three datasets, the method yields gains in sensitivity and accuracy over an XGBoost baseline; the sliding-window approach balances responsiveness and redundancy. Limitations include fixed completeness thresholds and window sizes, limited interpretability, and lack of prospective validation. Future work will prioritize external validation on multi-center ICU datasets, integration of explainable AI to enhance transparency, exploration of multimodal fusion, and real-world ICU deployment. This framework presents an efficient, applicable solution for early sepsis detection, supporting scalable integration into hospital systems and critical care workflows.

## 5 Acknowledgements

This work was supported by the Key Clinical Research Incubation Project (No.2024IITZDA02) of Southwest Hospital, Chongqing, China.

# 6 References

[1] Wu, D., Luo, X.: 'Robust Latent Factor Analysis for Precise Representation of High-Dimensional and Sparse Data', IEEE/CAA J. Automatica Sinica, 2021, 8, (4), pp. 796–805

[2] Wu, D., Luo, X., Shang, M., et al.: 'A Deep Latent Factor Model for High-Dimensional and Sparse Matrices in Recommender Systems', IEEE Trans. Syst. Man Cybern. Syst., 2021, 51, (7), pp. 4285–4296

[3] Wu, D., Luo, X., Shang, M., et al.: 'A Data Characteristic Aware Latent Factor Model for Web Services QoS Prediction', IEEE Trans. Knowl. Data Eng., 2022, 34, (6), pp. 2525–2538

[4] Wu, D., Shang, M., Luo, X., et al.: 'An L1 and L2-norm Oriented Latent Factor Model for Recommender Systems', IEEE Trans. Neural Netw. Learn. Syst., 2022, 33, (10), pp. 5775–5788

[5] Wu, D., He, Y., Luo, X., et al.: 'A Latent Factor Analysis-Based Approach to Online Sparse Streaming Feature Selection', IEEE Trans. Syst. Man Cybern. Syst., 2022, 52, (11), pp. 6744–6758

[6] Wu, D., Li, Z., Chen, F., et al.: 'Online Sparse Streaming Feature Selection With Gaussian Copula', IEEE Trans. Big Data, 2024, 10, (1), pp. 92–107

[7] Wu, D., Luo, X., He, Y., et al.: 'A Prediction-Sampling-Based Multilayer-Structured Latent Factor Model for Accurate Representation to High-Dimensional and Sparse Data', IEEE Trans. Neural Netw. Learn. Syst., 2024, 35, (3), pp. 3845–3858

[8] Wu, D., Hu, Y., Liu, K., et al.: 'An Outlier-Resilient Autoencoder for Representing High-Dimensional and Incomplete Data', IEEE Trans. Emerg. Topics Comput., 2024

[9] Chen, J., Wang, R., Wu, D., Luo, X.: 'A Differential Evolution-Enhanced Position-Transitional Approach to Latent Factor Analysis', IEEE Trans. Emerg. Topics Comput. Intell., 2023, 7, (2), pp. 389–401

[10] Wu, D., Li, Z., Yu, Z., et al.: 'Robust Low-Rank Latent Feature Analysis for Spatio-Temporal Signal Recovery', IEEE Trans. Neural Netw. Learn. Syst., 2023, 34, (11)

[11] Wu, D., He, Y., Luo, X.: 'A Graph-Incorporated Latent Factor Analysis Model for High-Dimensional and Sparse Data', IEEE Trans. Emerg. Topics Comput., 2023, 11, (4), pp. 907–917

[12] Wu, D., Pang, P., He, Y., Luo, X.: 'A Double-Space and Double-Norm Ensembled Latent Factor Model for Highly Accurate Web Service QoS Prediction', IEEE Trans. Serv. Comput., 2023, 16, (2), pp. 802–814

[13] Li, Z., Li, S., Luo, X.: 'A Novel Machine Learning System for Industrial Robot Arm Calibration', IEEE Trans. Circuits Syst. II: Express Briefs, 2023

[14] Qin, W., Luo, X., Zhou, M.: 'Adaptively-Accelerated Parallel Stochastic Gradient Descent for High-Dimensional and Incomplete Data Representation Learning', IEEE Trans. Big Data, 2024, 10, (1), pp. 92–107

[15] Chen, T., Li, S., Qiao, Y., Luo, X.: 'A Robust and Efficient Ensemble of Diversified Evolutionary Computing Algorithms for Accurate Robot Calibration', IEEE Trans. Instrum. Meas., 2024, 73, pp. 1–14

[16] Zeng, N., Li, X., Wu, P., Li, H., Luo, X.: 'A Novel Tensor Decomposition-Based Efficient Detector for Low-Altitude Aerial Objects With Knowledge Distillation Scheme', IEEE/CAA J. Automatica Sinica, 2024, 11, (2), pp. 487–501

[17] Yan, J., Jin, L., Luo, X., Li, S.: 'Modified RNN for Solving Comprehensive Sylvester Equation With TDOA Application', IEEE Trans. Neural Netw. Learn. Syst., 2023

[18] Chen, X., Luo, X., Jin, L., Li, S., Liu, M.: 'Growing Echo State Network With an Inverse-Free Weight Update Strategy', IEEE Trans. Cybern., 2023, 53, (2), pp. 753–764

[19] Luo, X., Wu, H., Wang, Z., Wang, J., Meng, D.: 'A Novel Approach to Large-Scale Dynamically Weighted Directed Network Representation', IEEE Trans. Pattern Anal. Mach. Intell., 2022, 44, (12), pp. 9756–9773

[20] Bi, F., He, T., Ong, Y. S., Luo, X.: 'Graph Linear Convolution Pooling for Learning in Incomplete High-Dimensional Data', IEEE Trans. Knowl. Data Eng., 2024

[21] Wu, H., Qiao, Y., Luo, X.: 'A Fine-Grained Regularization Scheme for Nonnegative Latent Factorization of High-Dimensional and Incomplete Tensors', IEEE Trans. Serv. Comput., 2024

[22] Zhou, Y., Liu, Z., Luo, X., Hu, L., Zhou, M.: 'Generalized Nesterov's Acceleration-Incorporated, Non-negative and Adaptive Latent Factor Analysis', IEEE Trans. Serv. Comput., 2022, 15, (5), pp. 2809–2823

[23] Luo, X., Chen, M., Wu, H., Liu, Z., Yuan, H., Zhou, M.: 'Adjusting Learning Depth in Non-negative Latent Factorization of Tensors for Accurately Modeling Temporal Patterns in Dynamic QoS Data', IEEE Trans. Autom. Sci. Eng., 2022, 18, (4), pp. 2142–2155

[24] Luo, X., Liu, Z., Jin, L., Zhou, Y., Zhou, M.: 'Symmetric Non-negative Matrix Factorization-Based Community Detection Models and Their Convergence Analysis', IEEE Trans. Neural Netw. Learn. Syst., 2022, 33, (3), pp. 1203–1215

[25] Li, J., Tan, F., He, C., Wang, Z., Song, H., Hu, P., Luo, X.: 'Saliency-Aware Dual Embedded Attention Network for Multivariate Time-Series Forecasting in Information Technology Operations', IEEE Trans. Ind. Informatics, 2024, 20, (3), pp. 4206–4217

[26] Luo, X., Zhou, Y., Liu, Z., Zhou, M.: 'Fast and Accurate Non-negative Latent Factor Analysis on High-Dimensional and Sparse Matrices in Recommender Systems', IEEE Trans. Knowl. Data Eng., 2023, 35, (4), pp. 3897–3911

[27] Luo, X., Zhong, Y., Wang, Z., Li, M.: 'An Alternating-Direction-Method of Multipliers-Incorporated Approach to Symmetric Non-negative Latent Factor Analysis', IEEE Trans. Neural Netw. Learn. Syst., 2023, 34, (8), pp. 4826–4840

[28] Luo, X., Wang, L., Hu, P., Hu, L.: 'Predicting Protein–Protein Interactions Using Sequence and Network Information via Variational Graph Autoencoder', IEEE/ACM Trans. Comput. Biol. Bioinform., 2023, 20, (5), pp. 3182–3194

[29] Luo, X., Chen, J., Yuan, Y., Wang, Z.: 'Pseudo Gradient-Adjusted Particle Swarm Optimization for Accurate Adaptive Latent Factor Analysis', IEEE Trans. Syst. Man Cybern. Syst., 2023

[30] Chen, J., Liu, K., Luo, X., Yuan, Y., Sedraoui, K., Al-Turki, Y., Zhou, M.: 'A State-Migration Particle Swarm Optimizer for Adaptive Latent Factor Analysis of High-Dimensional and Incomplete Data', IEEE/CAA J. Automatica Sinica, 2024

[31] Yang, W., Li, S., Luo, X.: 'Data-Driven Vibration Control: A Review', IEEE/CAA J. Automatica Sinica, 2024

[32] Liao, X., Hoang, K., Luo, X.: 'Local Search-Based Anytime Algorithms for Continuous Distributed Constraint Optimization Problems', IEEE/CAA J. Automatica Sinica, 2024

[33] Zhong, Y., Liu, K., Gao, S., Luo, X.: 'Alternating-Direction-Method of Multipliers-Based Adaptive Nonnegative Latent Factor Analysis', IEEE Trans. Emerg. Topics Comput., 2024

[34] Chen, T., Yang, W., Zhang, Z., Luo, X.: 'An Efficient Industrial Robot Calibrator With Multi-Planer Constraints', IEEE Trans. Ind. Informatics, 2024

[35] Yuan, Y., Wang, Y., Luo, X.: 'A Node-Collaboration-Informed Graph Convolutional Network for Highly-Accurate Representation to Undirected Weighted Graph', IEEE Trans. Neural Netw. Learn. Syst., 2024

[36] Lin, M., Liu, J., Chen, H., Xu, X., Luo, X., Xu, Z.: 'A 3D Convolution-Incorporated Dimension Preserved Decomposition Model for Traffic Data Prediction', IEEE Trans. Intell. Transp. Syst., 2024

[37] Luo, Z., Jin, X., Luo, Y., Zhou, Q., Luo, X.: 'Analysis of Students' Positive Emotion and Smile Intensity Using Sequence-Relative Key-Frame Labeling and Deep-Asymmetric Convolutional Neural Network', IEEE/CAA J. Automatica Sinica, 2024

[38] Huang, P., Luo, X.: 'FDTs: A Feature Disentangled Transformer for Interpretable Squamous Cell Carcinoma Grading', IEEE/CAA J. Automatica Sinica, 2024

[39] Chen, J., Yuan, Y., Luo, X.: 'SDGNN: Symmetry Preserving Dual Stream Graph Neural Networks', IEEE/CAA J. Automatica Sinica, 2024

[40] Yuan, Y., Li, J., and Luo, X.: 'A fuzzy PID-incorporated stochastic gradient descent algorithm for fast and accurate latent factor analysis', IEEE Trans. Fuzzy Syst., 2024, 32, (2), pp. 445–459

[41] He, Z., Lin, M., and Luo, X.: 'Structure-preserved self-attention for fusion image information in multiple color spaces', IEEE Trans. Neural Netw. Learn. Syst., 2024, 35, (11), pp. 8778–8790

[42] Huang, P., and Luo, X.: 'FDTs: A feature disentangled transformer for interpretable squamous cell carcinoma grading', IEEE/CAA J. Autom. Sinica, 2024, 11, (2), pp. 487–501

[43] Yuan, Y., Wang, Y., and Luo, X.: 'A node-collaboration-informed graph convolutional network for highly-accurate representation to undirected weighted graph', IEEE Trans. Neural Netw. Learn. Syst., 2024, 35, (3), pp. 3845–3858

[44] Wu, Q., Ye, F., Gu, Q., et al.: 'A Customised Down-Sampling Machine Learning Approach for Sepsis Prediction', Int. J. Med. Inform., 2024, 184, 105365

[45] Delahanty, R. J., Alvarez, J. A., Flynn, L. M., et al.: 'Development and Evaluation of a Machine Learning Model for the Early Identification of Patients at Risk for Sepsis', Ann. Emerg. Med., 2019, 73, (3), pp. 334–344

[46] Bloch, E., Rotem, T., Cohen, J., et al.: 'Machine Learning Models for Analysis of Vital Signs Dynamics: A Case for Sepsis Onset Prediction', J. Healthc. Eng., 2019, 2019, Article 5930379

[47] Bedoya, A. D., Futoma, J., Clement, M. E., et al.: 'Machine Learning for Early Detection of Sepsis: An Internal and Temporal Validation Study', JAMIA Open, 2020, 3, pp. 252–260

[48] J. Futoma, S. Hariharan, K. Heller, et al., "An improved multi-output Gaussian process RNN for the early detection of sepsis," Machine Learning for Healthcare, 2017, pp. 243–254.

[49] T. Desautels, M. Calvert, P. Hoffman, et al., "Prediction of sepsis in the intensive care unit with minimal electronic health record data: a machine learning approach," JMIR Medical Informatics, 2016, 4(3): e28.

[50] D. Celi, L. Ghassemi, F. Naumann, et al., "ICU data collaboration for sepsis prediction and early warning systems," Nature Medicine, 2019, 25(1), pp. 44–50.

[51] S. Moor, C. R. Wulff, P. Svensson, et al., "Early prediction of sepsis in the ICU using machine learning: a systematic review," Critical Care Medicine, 2021, 49(12), pp. e1221–e1232.

[52] M. Reyna, C. Josef, S. Jeter, et al., "Early prediction of sepsis from clinical data: the PhysioNet/Computing in Cardiology Challenge 2019," Critical Care Medicine, 2020, 48(2), pp. 210–217.

[53] J. Nemati, M. Holder, A. Razmi, et al., "An interpretable machine learning model for accurate prediction of sepsis in the ICU," Critical Care Explorations, 2021, 3(1): e0331.

[54] A. Kamaleswaran, M. Akbilgic, "Artificial intelligence approaches for sepsis: An updated review," Frontiers in Digital Health, 2022, 4: 867107.

[55] J. van Wyk, E. Khojandi, D. Davis, "Early prediction of sepsis in the ICU using machine learning: A systematic survey," Computers in Biology and Medicine, 2019, 115: 103375.

[56] A. Schamoni, T. Frank, F. Fiedler, et al., "Sepsis prediction in intensive care using machine learning and temporal data," Artificial Intelligence in Medicine, 2021, 118: 102104.

[57] M. Scherpf, C. Gräßer, T. Malberg, et al., "Predicting sepsis with a recurrent neural network using electronic health records," Computers in Biology and Medicine, 2019, 113: 103395.

[58] Y. Bengio, A. Courville, P. Vincent, "Representation learning: A review and new perspectives," IEEE Trans. Pattern Anal. Mach. Intell., 2013, 35(8), pp. 1798–1828.

[59] K. Baldi, "Autoencoders, unsupervised learning, and deep architectures," ICML Workshop on Unsupervised Feature Learning, 2012, pp. 37–49.

[60] D. Erhan, Y. Bengio, A. Courville, et al., "Why does unsupervised pre-training help deep learning?," Journal of Machine Learning Research, 2010, 11, pp. 625–660.

[61] I. Goodfellow, Y. Bengio, A. Courville, Deep Learning, MIT Press, 2016.

[62] C. Doersch, "Tutorial on variational autoencoders," arXiv preprint, arXiv:1606.05908, 2016.

[63] A. Yoon, J. Zame, M. van der Schaar, "Estimating missing data in temporal data streams using multi-directional recurrent neural networks," IEEE Trans. Biomed. Eng., 2019, 66(5), pp. 1477–1490.

[64] M. Lipton, D. Kale, C. Wetzel, et al., "Directly modeling missing data in sequences with RNNs: Improved classification of clinical time series," Machine Learning for Healthcare, 2016, pp. 253–270.

[65] E. Choi, M. T. Bahadori, L. Song, W. Stewart, J. Sun, "GRAM: Graph-based attention model for healthcare representation learning," KDD, 2017, pp. 787–795.

[66] P. Che, Z. Cheng, J. Sun, "Recurrent neural networks for multivariate time series with missing values," Scientific Reports, 2018, 8, 6085.

[67] J. Yoon, W. R. Zame, M. van der Schaar, "GAIN: Missing data imputation using generative adversarial nets," ICML, 2018, pp. 5689–5698.